\documentclass[lettersize,journal]{IEEEtran}
\usepackage{amsmath,amsfonts,amssymb}
\usepackage{algorithm}
\usepackage{array}
\usepackage[caption=false,font=normalsize,labelfont=sf,textfont=sf]{subfig}
\usepackage{textcomp}
\usepackage{stfloats}
\usepackage{url}
\usepackage{verbatim}
\usepackage{graphicx}
\usepackage{cite}
\usepackage{xcolor}
\usepackage{svg}

\usepackage{algpseudocode}
\usepackage{braket}
\usepackage{booktabs}
\usepackage{tabularx}
\usepackage{array}
\usepackage{tikz}
\usepackage[normalem]{ulem}

\begin{document}

\title{Quantum Closed Loop Inverse Kinematics Through\\ Hilbert-Space Encoding of Robot Kinematics}

\author{Andrea Morghen,
Pierluigi Arpenti,
Roberto Schiattarella,
Giovanni Acampora,
and Bruno Siciliano%
\thanks{The authors are with the University of Naples Federico II, Naples, Italy.}%
}

\markboth{Preprint}%
{Morghen \MakeLowercase{\textit{et al.}}:
Quantum Closed Loop Inverse Kinematics Through
Hilbert-Space Encoding of Robot Kinematics}

\maketitle

\begin{abstract}

The integration of quantum computing into robotics requires mathematical representations that preserve the geometric structure and configuration dependence of robot motion. This paper introduces a structure-preserving quantum representation of rigid-body transformations and serial-manipulator kinematics, establishing a correspondence between classical kinematic models and parameterized quantum circuits. Building on the Bloch-sphere representation, spatial directions are encoded into single-qubit states, while quantum rotations reproduce the ordered composition of reference-frame transformations. Translational contributions are combined through quantum-state aggregation, with their metric scales retained classically, enabling the reconstruction of end-effector position and orientation from Pauli expectation values. The formulation preserves the original joint variables and accommodates different kinematic conventions and joint types. Beyond forward kinematics, we establish that the manipulator geometric Jacobian can be reconstructed from quantum-observable derivatives using the parameter-shift rule, with analytical equivalence to its classical counterpart under ideal readout. To demonstrate the applicability of this representation to robot-level algorithms, we introduce Quantum Closed Loop Inverse Kinematics, a hybrid closed-loop method that exploits quantum-derived pose and Jacobian information. The proposed framework provides a geometrically interpretable foundation for developing and investigating quantum and hybrid computational methods for robotic kinematics and motion.

\end{abstract}

\section{Introduction}
\label{sec:introduction}

\IEEEPARstart{R}{obotics} grounds many of its core methodologies in mathematical models that relate the internal configuration of a robot to geometry and motion in physical space. For serial manipulators, kinematics provides this fundamental relation by mapping joint variables to the poses of the links and end effector, while differential kinematics relates joint rates to task-space velocities. These mappings constitute a common computational basis for workspace and singularity analysis, inverse kinematics, trajectory generation, motion planning, task-space control, interaction control, and sensor-based methods such as visual servoing. A consistent kinematic representation is therefore not limited to pose evaluation, but supports several layers of manipulator modelling, planning, and control.
For serial manipulators, kinematics is naturally expressed through an ordered composition of rigid-body transformations between consecutive link frames. The resulting forward kinematic map determines the pose of each frame, and in particular the end-effector pose, as a function of the joint configuration. Its differential counterpart is characterized by the manipulator Jacobian, which locally relates joint-space variations to task-space motion \cite{Craig2005IntroductionRobotics,Siciliano2009Robotics}.

Robotic algorithms have historically been developed within the paradigm of classical digital computation, enabling increasingly complex pipelines for modelling, optimisation, estimation, planning, control, and learning. Quantum computing introduces a different computational substrate based on information-processing principles that depart from classical bits and deterministic numerical representations. Established theoretical quantum advantages concern specific computational problems, including integer factorisation and discrete logarithms through Shor's algorithm, unstructured search through Grover's algorithm, and the simulation of quantum physical systems \cite{Feynman1982Simulating,Lloyd1996UniversalQuantumSimulators,Shor1994Algorithms,Grover1996Fast}. At the same time, currently available noisy intermediate-scale quantum (NISQ) devices remain constrained by limited qubit counts, decoherence, gate errors, finite sampling, and compilation overhead, restricting the depth and reliability of executable circuits \cite{Preskill2018NISQ,Bharti2022NISQAlgorithms}. These limitations motivate hybrid quantum--classical approaches in which quantum processing is assigned to specific representations or computational primitives while the remaining operations are retained classically.

The relevance of these developments to robotics is still largely open. Recent works have started to investigate quantum and hybrid quantum--classical methods for robot modelling, navigation, swarm coordination, posture representation, planning, learning, and optimisation \cite{Yan2024QuantumRobotics,Hohenfeld2024QuantumNavigation,Chella2023QuantumSwarm,Mannone2025QuantumSwarm,Otani2025Posture}. However, progress in quantum robotics cannot rely only on reformulating final task objectives as generic quantum optimisation problems. It also requires representations that preserve the geometric structure and configuration dependence of the underlying robotic system. For robotics applications, quantum states, gates, measurements, and observables should therefore remain connected to interpretable quantities such as joint variables, frame directions, link displacements, task-space positions, end-effector orientations, and their variations with respect to the robot configuration.
This paper takes this requirement as its starting point. The objective is not to replace conventional kinematic algorithms, nor to claim a present computational quantum advantage over classical robotic computation. Instead, we develop a geometrically interpretable quantum representation of rigid-body and serial-manipulator kinematics in which the dependence on the robot configuration is retained explicitly and the resulting quantum readouts remain directly associated with task-space pose quantities. Crucially, this dependence is preserved at the differential level, enabling the same representation to provide access to local kinematic information required by downstream robotic algorithms.

A direct route toward this goal is provided by the Bloch-sphere representation of a single qubit. The expectation values of the three Pauli operators define a real three-dimensional Bloch vector, which has unit norm for pure states. Parameterised single-qubit unitaries act as rotations of this vector, establishing a precise connection between quantum-state evolution and rotations in three-dimensional Euclidean space \cite{Nielsen2010QuantumComputation}. This connection is particularly natural for serial-manipulator kinematics, where link directions and frame axes are propagated through ordered products of spatial rotations. This correspondence provides the geometric basis for the quantum representation developed in this work. Initial studies have explored the use of qubit states and observables to represent specific aspects of robot kinematics, including orientations, link postures, and end-effector quantities, typically through application-specific formulations and individual manipulator case studies \cite{Zioui2021Orientation,Fazilat2022ForwardKinematics,Otani2025Posture}. These works demonstrate the feasibility of representing robotic geometric quantities through quantum states. However, to the best of our knowledge, existing approaches do not provide a unified constructive framework that connects quantum representations of rigid-body transformations with their ordered composition along configuration-dependent serial kinematic chains and with the differential information required to recover local task-space motion. 
This work addresses this gap by developing a bottom-up quantum representation of rigid-body kinematics and extending it to configuration-dependent serial manipulators and their differential kinematics. The main contributions are:
\begin{itemize}
\item A bottom-up quantum representation of rigid-body transformations between reference frames and their ordered composition along configuration-dependent serial kinematic chains, preserving the correspondence between quantum readouts and robot pose quantities.

\item A reconstruction of the manipulator geometric Jacobian from parameter-shift derivatives of the quantum pose readouts, together with an analytical proof of equivalence with the conventional geometric Jacobian under ideal expectation-value readout.

\item Quantum Loop Inverse Kinematics (\textsc{QLIK}), a hybrid quantum--classical closed-loop inverse-kinematics scheme that uses the proposed quantum representation to evaluate pose and Jacobian information.

\item An experimental campaign comprising validation of the quantum kinematic representation across multiple manipulators under ideal simulation and quantum-hardware tests to assess its applicability on current devices, together with an evaluation of \textsc{QLIK} through dedicated inverse-kinematics experiments on a physical manipulator.
\end{itemize}
The contribution of this work is primarily methodological. Rather than replacing established classical kinematic algorithms or claiming a computational quantum advantage, the proposed framework provides a geometrically interpretable interface between rigid-body kinematics and quantum computation. The representation preserves configuration-dependent pose information and enables the corresponding geometric Jacobian to be recovered from quantum readout derivatives, while \textsc{QLIK} demonstrates how this interface can be embedded directly into a robot-level inverse-kinematics algorithm.

The remainder of this article is organized as follows. Section~\ref{sec:related_work} reviews related work on quantum representations and algorithms for robotic kinematics. Section~\ref{sec:methodological_background} introduces the classical and quantum concepts required by the formulation. 

\section{Related Work}
\label{sec:related_work}
Research on quantum computing for robotic manipulation can be broadly divided into two methodological levels. The first introduces quantum, hybrid quantum--classical, or quantum-inspired algorithms as solvers for search, control, inverse kinematics, or trajectory optimisation. The second seeks to represent the kinematic quantities themselves through quantum states, parameterised unitaries, and observable expectation values. This distinction is important: encoding candidate joint configurations for optimisation does not imply that the kinematic map relating those configurations to task-space quantities is itself represented within the quantum circuit. The following review therefore separates solver-level quantum optimisation from quantum-state representations of manipulator kinematics, and positions the proposed rigid-body and serial-kinematic representation with respect to both.

\subsection{Quantum Optimisation for Robotic Manipulators}
\label{subsec:rw_quantum_optimisation}

Quantum and hybrid approaches to robotic manipulation have mainly introduced quantum computation at the solver level, by casting control or inverse kinematics as search and optimisation problems. Dahassa and Zioui formulated optimal control for a six-jointed manipulator as a Grover-based search over discretised control inputs, using a classical controller as a pseudo-oracle due to the impracticality of constructing the full quantum oracle \cite{Dahassa2025GroverArm}. Salloum et al.\ mapped planar inverse kinematics to a QUBO problem through joint-angle discretisation and one-hot constraints, solved it on D-Wave annealing hardware, and evaluated the decoded configurations using classical forward kinematics \cite{Salloum2026AnnealingIK}. Nigatu et al.\ combined a parameterised quantum circuit trained as a forward-kinematics surrogate with Grover-based search over discretised configurations and design parameters \cite{Nigatu2025QMLGrover}. Quantum-inspired IK methods based on neural-network variants or quantum-behaved metaheuristics are related but distinct, since they do not require quantum-circuit execution \cite{Fazilat2025QNNIK,Dereli2020QPSOIK}.
Overall, these approaches use quantum resources primarily for search, discrete optimisation, or learned function approximation, rather than deriving an analytic, continuous, joint-parameterised quantum representation of the manipulator kinematics from which both pose quantities and their local differential information can be recovered.
\begin{table*}[!t]
\centering
\caption{Comparison with representative quantum, hybrid, and quantum-inspired approaches to robotic kinematics and optimisation.}
\label{tab:related_work_comparison_compact}

\renewcommand{\arraystretch}{1.12}
\setlength{\tabcolsep}{2.5pt}
\scriptsize

\begin{tabularx}{\textwidth}{
@{}
>{\raggedright\arraybackslash}p{2.1cm}
>{\raggedright\arraybackslash}p{2.5cm}
>{\raggedright\arraybackslash}p{4.0cm}
>{\raggedright\arraybackslash}p{4.0cm}
>{\raggedright\arraybackslash}X
@{}
}
\toprule

\textbf{Work} &
\textbf{Quantum role} &
\textbf{Model construction} &
\textbf{Pose formation and readout} &
\textbf{Differential information}
\\

\midrule

Dahassa and Zioui~\cite{Dahassa2025GroverArm} &
Grover-based control search &
Discrete control search with classical pseudo-oracle. &
Measured control codes with classical propagation. &
Not represented.
\\

Salloum et al.~\cite{Salloum2026AnnealingIK} &
Quantum-annealing IK &
QUBO from discretised joint angles. &
Bitstrings decoded and evaluated through classical FK. &
Not represented.
\\

Nigatu et al.~\cite{Nigatu2025QMLGrover} &
QML-assisted search &
Learned parameterised quantum-circuit FK surrogate. &
Candidate configurations with classical verification. &
Not analytically derived.
\\

Zioui et al.~\cite{Zioui2021Orientation} &
Analytic geometric encoding &
Single-qubit representation of spatial rotations. &
Orientation from Pauli-observable expectation values. &
Not derived.
\\

Fazilat et al.~\cite{Fazilat2022ForwardKinematics} &
Quantum--classical FK &
Robot-specific quaternion--Pauli formulation. &
Quantum rotations with classical translational composition. &
Not derived.
\\

Otani et al.~\cite{Otani2025Posture} &
Link-posture encoding and hybrid IK &
Parameterized qubit representation of link directions. &
Pauli readout with classical length weighting and summation. &
Not derived from readout derivatives.
\\

\textbf{This work} &
\textbf{Quantum kinematic representation and hybrid IK} &
\textbf{Rigid-body transformations composed along a joint-parameterised serial chain.} &
\textbf{Observable-based terminal pose representation.} &
\textbf{Geometric Jacobian reconstructed from parameter-shift readout derivatives.}
\\

\bottomrule
\end{tabularx}
\end{table*}

\subsection{Quantum Representations of Manipulator Kinematics}
\label{subsec:rw_quantum_kinematics}
A more limited line of work has investigated whether robot geometry can be represented directly through quantum states and observables. Zioui et al.\ related robotic-arm and rigid-body orientation to single-qubit rotations by exploiting the correspondence between Bloch vectors, Pauli operators, and spatial rotations \cite{Zioui2021Orientation}. The approach was demonstrated on SCARA and articulated manipulators, but remains primarily orientational and does not develop a representation of complete rigid-body transformations and their configuration-dependent composition along a general serial kinematic chain.
Fazilat et al.\ extended this idea to the forward kinematics of an ABB IRB~140 manipulator \cite{Fazilat2022ForwardKinematics}. They used quaternion--Pauli correspondences to represent elementary kinematic transformations and recover end-effector position and orientation. However, translational composition is still performed through classical arithmetic, and the method does not establish a general quantum composition rule for configuration-dependent rigid-body transformations or derive the corresponding manipulator Jacobian from quantum readouts.
Otani et al.\ encoded robot-link directions in qubit states and reconstructed the end-effector position from Pauli expectation values \cite{Otani2025Posture}. Their formulation is developed for specific link geometries and does not provide a general construction from arbitrary relative rigid-body transformations. Moreover, no configuration-wise equivalence with classical forward kinematics is established, while the reported inverse-kinematics experiments retain large residual position errors, on the order of one metre even for the entangled formulation. Their extension based on two-qubit gates further departs from an explicit rigid-body composition rule, since no analytical equivalence between these operations and the corresponding classical kinematics is derived.

Taken together, these studies demonstrate that quantum states and observables can represent selected geometric quantities relevant to robotic manipulation, but they do not establish, within a unified construction, a quantum representation of complete rigid-body transformations, their ordered composition along configuration-dependent serial kinematic chains, and the recovery of the manipulator geometric Jacobian from derivatives of the resulting quantum readouts. The present work addresses this gap by developing such a representation and by using the resulting pose and Jacobian information within the proposed \textsc{QLIK} closed-loop inverse-kinematics scheme. Table~\ref{tab:related_work_comparison_compact} summarises the comparison along the dimensions most relevant to the proposed formulation.


\section{Methodological Background}
\label{sec:methodological_background}

This section introduces the classical kinematic and quantum-mechanical concepts required by the proposed formulation. We first establish the notation for serial-manipulator pose and differential kinematics, with particular emphasis on the propagation of frame axes along an ordered kinematic chain. We then recall the single-qubit Bloch-vector representation, its correspondence with three-dimensional rotations, the reduced-state linearity used for weighted spatial aggregation, and the parameter-shift rule employed for differential readout.

\subsection{Serial-Manipulator Kinematics}
\label{subsec:background_robot_pose}

Consider a serial open-chain manipulator with joint configuration \(\mathbf q=[q_1,\ldots,q_n]^{\mathsf T}\in\mathcal Q\), where \(\mathcal Q=\prod_{j=1}^{n}\mathcal Q_j\subseteq\mathbb R^n\) and \(\mathcal Q_j=[q_j^{-},q_j^{+}]\) is the admissible interval of the \(j\)-th revolute or prismatic coordinate. Let \(\mathcal F_B\) denote the fixed-base frame and let \(\mathcal F_0,\ldots,\mathcal F_n\) be an ordered sequence of frames associated with the kinematic chain, with \(\mathcal F_n\) identified with the terminal frame. A rigid-body transformation is denoted by \(\mathbf T_b^{a}=(\mathbf R_b^{a},\mathbf p_b^{a})\in SE(3)\), where \(\mathbf R_b^{a}\in SO(3)\) describes the orientation of \(\mathcal F_b\) relative to \(\mathcal F_a\), and \(\mathbf p_b^{a}\in\mathbb R^3\) is the position of the origin of \(\mathcal F_b\) expressed in \(\mathcal F_a\). Rigid-body transformations compose according to \((\mathbf R,\mathbf p)(\mathbf R',\mathbf p')=(\mathbf R\mathbf R',\mathbf p+\mathbf R\mathbf p')\).
Let \(\mathbf T_0=\mathbf T_0^{B}=(\mathbf R_0,\mathbf p_0)\) denote the fixed pose of \(\mathcal F_0\) in \(\mathcal F_B\) and let \(\mathbf T_i^{i-1}(\mathbf q)\in SE(3)\), \(i=1,\ldots,n\), denote the configuration-dependent transformations between consecutive frames. The terminal forward-kinematics map is
\begin{equation}
\mathbf T_n(\mathbf q)=\mathbf T_0\prod_{i=1}^{n}\mathbf T_i^{i-1}(\mathbf q)=\left(\mathbf R_n(\mathbf q),\mathbf p_n(\mathbf q)\right),
\label{eq:background_terminal_pose}
\end{equation}
where the multiplication order follows the frame sequence. Throughout the paper, omission of the right superscript indicates that the corresponding quantity is expressed in \(\mathcal F_B\). For \(\alpha\in\{x,y,z\}\), let \(\mathbf e_\alpha\) denote the corresponding canonical basis vector. The terminal-frame axes expressed in the base frame are defined as \(\mathbf r_{n,\alpha}(\mathbf q):=\mathbf R_n(\mathbf q)\mathbf e_\alpha\), and therefore
$\mathbf R_n(\mathbf q)=
\begin{bmatrix}
\mathbf r_{n,x}(\mathbf q) &
\mathbf r_{n,y}(\mathbf q) &
\mathbf r_{n,z}(\mathbf q)
\end{bmatrix}$.
Because \(\mathbf R_n(\mathbf q)\in SO(3)\), the three axes form a right-handed orthonormal basis and satisfy \(\mathbf r_{n,y}(\mathbf q)=\mathbf r_{n,z}(\mathbf q)\times\mathbf r_{n,x}(\mathbf q)\).
The terminal differential kinematics are described by the geometric Jacobian
\[
\begin{bmatrix}
\dot{\mathbf p}_n\\
\boldsymbol{\omega}_n
\end{bmatrix}
=
\mathbf J_G(\mathbf q)\dot{\mathbf q},
\qquad
\mathbf J_G(\mathbf q)=
\begin{bmatrix}
\mathbf J_v(\mathbf q)\\
\mathbf J_{\omega}(\mathbf q)
\end{bmatrix}
\in\mathbb R^{6\times n},
\]
where \(\mathbf J_v(\mathbf q)\in\mathbb R^{3\times n}\) and \(\mathbf J_{\omega}(\mathbf q)\in\mathbb R^{3\times n}\) are the translational and angular blocks, respectively, and both \(\dot{\mathbf p}_n\) and \(\boldsymbol{\omega}_n\) are expressed in \(\mathcal F_B\). Since \(\mathbf p_n=\mathbf p_n(\mathbf q)\), the translational block satisfies \(\mathbf J_v(\mathbf q)=\partial\mathbf p_n(\mathbf q)/\partial\mathbf q\).
For any vector \(\mathbf v=[v_x,v_y,v_z]^{\mathsf T}\in\mathbb R^3\), define
\[
[\mathbf v]_{\times}:=
\begin{bmatrix}
0 & -v_z & v_y\\
v_z & 0 & -v_x\\
-v_y & v_x & 0
\end{bmatrix}.
\]
This matrix satisfies \([\mathbf v]_{\times}\mathbf w=\mathbf v\times\mathbf w\), \([\mathbf v]_{\times}^{\mathsf T}=-[\mathbf v]_{\times}\), and \([\mathbf v]_{\times}^{\mathsf T}[\mathbf v]_{\times}=\|\mathbf v\|_2^2\mathbf I_3-\mathbf v\mathbf v^{\mathsf T}\). Let \(\mathbf J_{\omega}^{(j)}(\mathbf q)\) denote the \(j\)-th column of \(\mathbf J_{\omega}(\mathbf q)\). The derivative of each terminal-frame axis satisfies
\begin{equation}
\frac{\partial\mathbf r_{n,\alpha}(\mathbf q)}{\partial q_j}
=
\mathbf J_{\omega}^{(j)}(\mathbf q)\times\mathbf r_{n,\alpha}(\mathbf q)
=
-[\mathbf r_{n,\alpha}(\mathbf q)]_{\times}\mathbf J_{\omega}^{(j)}(\mathbf q).
\label{eq:background_terminal_axis_derivative}
\end{equation}
Inverse kinematics seeks a configuration \(\mathbf q^\star\in\mathcal Q\) whose terminal pose satisfies a prescribed task-space target, whereas differential inverse-kinematics methods iteratively update the joint configuration through the geometric Jacobian.

\subsection{Single-Qubit Spatial Encoding and Rotations}
\label{subsec:qubit_spatial_directions}

A qubit is a two-level quantum system whose state is described by a density operator \(\rho\in\mathcal D(\mathbb C^2)\), where \(\mathcal D(\mathbb C^2)=\{\rho\in\mathbb C^{2\times2}\mid\rho\succeq0,\operatorname{Tr}(\rho)=1\}\). The two computational basis states are denoted by \(\lvert0\rangle=[1,0]^{\mathsf T}\) and \(\lvert1\rangle=[0,1]^{\mathsf T}\). A pure qubit state can be written as a normalized superposition \(\lvert\psi\rangle=c_0\lvert0\rangle+c_1\lvert1\rangle\), with \(|c_0|^2+|c_1|^2=1\), and is represented by the density operator \(\rho=\lvert\psi\rangle\langle\psi\rvert\). For \(\alpha\in\{x,y,z\}\), let \(\sigma_\alpha\) denote the corresponding Pauli observable,
\[
\sigma_x=
\begin{bmatrix}
0 & 1\\
1 & 0
\end{bmatrix},
\qquad
\sigma_y=
\begin{bmatrix}
0 & -\mathrm i\\
\mathrm i & 0
\end{bmatrix},
\qquad
\sigma_z=
\begin{bmatrix}
1 & 0\\
0 & -1
\end{bmatrix}.
\]
Every single-qubit state admits the Bloch decomposition \(\rho=\frac{1}{2}\left(\mathbf I_2+\mathbf r(\rho)\cdot\boldsymbol{\sigma}\right)\), where \(\boldsymbol{\sigma}:=(\sigma_x,\sigma_y,\sigma_z)\) and each Bloch component is the expectation value of the corresponding Pauli observable, \(\langle\sigma_\alpha\rangle_\rho=\operatorname{Tr}(\rho\sigma_\alpha)\). Hence,
\begin{equation}
\mathbf r(\rho):=
\begin{bmatrix}
\langle\sigma_x\rangle_\rho &
\langle\sigma_y\rangle_\rho &
\langle\sigma_z\rangle_\rho
\end{bmatrix}^{\mathsf T},
\qquad
\|\mathbf r(\rho)\|_2\leq1.
\label{eq:bloch_vector}
\end{equation}
In particular, the computational basis states correspond to the Bloch vectors \(\mathbf r(\lvert0\rangle\langle0\rvert)=\mathbf e_z\) and \(\mathbf r(\lvert1\rangle\langle1\rvert)=-\mathbf e_z\). The vector \(\mathbf r(\rho)\) geometrically represents the qubit state inside the unit Bloch ball; pure states lie on its boundary, the Bloch sphere \(\mathbb S^2\), and can therefore be associated with unit spatial directions. Accordingly, a direction \(\mathbf u\in\mathbb S^2\) is encoded by $\rho(\mathbf u):=\frac{1}{2}\left(\mathbf I_2+\mathbf u\cdot\boldsymbol{\sigma}\right),
$ so $\mathbf r\left(\rho(\mathbf u)\right)=\mathbf u$.
Thus, the three Cartesian components of the encoded direction correspond directly to the expectation values of the three Pauli observables. 
Spatial rotations admit an equivalent action on this representation. A rotation of angle \(\phi\) about the Cartesian axis \(\alpha\) is represented by the single-qubit unitary
\(
U_\alpha(\phi):=\exp\left(-\mathrm i\frac{\phi}{2}\sigma_\alpha\right)\in SU(2),
\)
whose action on the density operator induces the corresponding rotation of its Bloch vector,
\begin{equation}
\mathbf r\left(U_\alpha(\phi)\rho U_\alpha^\dagger(\phi)\right)
=
\mathbf R_\alpha(\phi)\mathbf r(\rho),
\label{eq:bloch_induced_rotation}
\end{equation}
where \(\mathbf R_\alpha(\phi)\in SO(3)\) is the associated right-handed spatial rotation. More generally, every \(\mathbf A\in SO(3)\) admits an \(SU(2)\) representative \(U_{\mathbf A}\), defined up to a global sign, whose conjugation action induces the corresponding rotation of the Bloch vector \cite{Nielsen2010QuantumComputation}. For an ordered composition \(\mathbf A=
\mathbf R_{1}\cdots
\mathbf R_{m}\), the corresponding unitary composition \(U_{\mathbf R}=U_{\mathbf R_1}\cdots U_{\mathbf R_m}\) preserves the same order, with the rightmost operation acting first.
The components of \(\mathbf r(\rho)\) are experimentally estimated from repeated measurements of the corresponding Pauli observables. While the theoretical development assumes exact expectation values, a finite number \(N_\alpha\) of independent measurements yields an estimator with variance
\(
\operatorname{Var}\left[\widehat r_\alpha\right]
=
\frac{1-r_\alpha^2(\rho)}{N_\alpha}
\leq
\frac{1}{N_\alpha},
\)
so that finite-shot readout introduces a sampling uncertainty that decreases with the number of measurements.

\subsection{Reduced-State Aggregation and Linearity of Pauli Readout}
\label{subsec:reduced_states}

Consider an auxiliary selector subsystem \(S\) and a single-qubit readout subsystem \(Q\). Let \(\mathcal L\) denote a finite set of selector branches, where each branch \(k\in\mathcal L\) is associated with a single-qubit state \(\rho_k\in\mathcal D(\mathbb C^2)\) and a probability \(\lambda_k\geq0\), with $\sum_{k\in\mathcal L}\lambda_k=1$.
For the selector-controlled preparation considered here, tracing out the selector subsystem yields the reduced readout state
\[
\rho_Q
=
\operatorname{Tr}_S(\rho_{SQ})
=
\sum_{k\in\mathcal L}\lambda_k\rho_k,
\]
where \(\rho_{SQ}\) denotes the joint selector--readout state.
For any single-qubit observable \(O\), linearity of the trace gives
\[
\langle O\rangle_{\rho_Q}
=
\operatorname{Tr}(\rho_Q O)
=
\sum_{k\in\mathcal L}
\lambda_k
\langle O\rangle_{\rho_k}.
\]
Applying this relation to the three Pauli observables and using the Bloch map defined in Eq.~\eqref{eq:bloch_vector} yields
\begin{equation}
\mathbf r(\rho_Q)
=
\sum_{k\in\mathcal L}
\lambda_k\mathbf r(\rho_k).
\label{eq:reduced_state_bloch_linearity}
\end{equation}
Hence, the Bloch vector of the readout qubit is the probability-weighted average of the branch Bloch vectors. Even when all branch states are pure, \(\rho_Q\) is generally mixed and its Bloch vector lies inside the Bloch ball. Equation~\eqref{eq:reduced_state_bloch_linearity} follows from reduced-state and expectation-value linearity; it does not represent a coherent vector addition within a single qubit.

\subsection{Parameter-Shift Differentiation}
\label{subsec:background_parameter_shift}

Consider an expectation value \(f(\mathbf x)\) produced by a parameterized quantum circuit, where \(\mathbf x=[x_1,\ldots,x_m]^{\mathsf T}\) denotes a generic vector of external parameters. Let \(\phi_g(\mathbf x)\) be the angle of a Pauli-generated gate occurrence
\(
U_g(\phi_g)
=
\exp\left(
-\mathrm i\frac{\phi_g}{2}P_g
\right),
\)
where \(P_g\) is a Pauli operator. For this generator, the derivative with respect to the gate angle is given exactly by the parameter-shift rule \cite{Li2017HybridQuantumClassical},
\[
\frac{\partial f}{\partial\phi_g}
=
\frac{1}{2}
\left[
f^{(g,+)}
-
f^{(g,-)}
\right],
\]
where \(f^{(g,\pm)}\) is obtained by shifting only the occurrence \(g\) as
\(\phi_g\mapsto\phi_g\pm\pi/2\), while all other circuit parameters remain unchanged.
Let \(\mathcal G_j\) denote the set of individual parameterized gate occurrences whose angles depend on \(x_j\). The classical chain rule then gives
\begin{equation}
\frac{\partial f(\mathbf x)}{\partial x_j}
=
\frac{1}{2}
\sum_{g\in\mathcal G_j}
\frac{\partial\phi_g(\mathbf x)}{\partial x_j}
\left[
f^{(g,+)}(\mathbf x)
-
f^{(g,-)}(\mathbf x)
\right].
\label{eq:background_parameter_shift}
\end{equation}
If the same parameter affects multiple elementary gate occurrences in the implemented circuit decomposition, each occurrence is shifted independently and the corresponding chain-rule contributions are summed according to Eq.~\eqref{eq:background_parameter_shift}. For an affine dependence \(\phi_g(\mathbf x)=c_gx_j+\phi_{g,0}\), the chain-rule factor reduces to \(\partial\phi_g/\partial x_j=c_g\). For a vector-valued readout constructed from multiple expectation values, Eq.~\eqref{eq:background_parameter_shift} is applied independently to each measured component. The resulting derivative is exact under exact expectation-value evaluation, whereas finite-shot execution introduces statistical uncertainty through the shifted expectation-value estimates.

\bibliographystyle{IEEEtran}
\bibliography{bibliography}

@book{Craig2005IntroductionRobotics,
  author    = {Craig, John J.},
  title     = {Introduction to Robotics: Mechanics and Control},
  edition   = {3},
  publisher = {Pearson Prentice Hall},
  year      = {2005}
}

@book{Siciliano2009Robotics,
  author    = {Siciliano, Bruno and Sciavicco, Lorenzo and Villani, Luigi and Oriolo, Giuseppe},
  title     = {Robotics: Modelling, Planning and Control},
  publisher = {Springer},
  year      = {2009},
  doi       = {10.1007/978-1-84628-642-1}
}

@article{Yan2024QuantumRobotics,
author = {Yan, Fei and Iliyasu, Abdullah M. and Li, Nianqiao and Salama, Ahmed S. and Hirota, Kaoru},
title = {Quantum Robotics: A Review of Emerging Trends},
journal = {Quantum Machine Intelligence},
volume = {6},
number = {2},
pages = {86},
year = {2024},
doi = {10.1007/s42484-024-00225-5}
}

@article{Chella2023QuantumSwarm,
  author  = {Chella, Antonio and Gaglio, Salvatore and Mannone, Maria and Pilato, Giovanni and Seidita, Valeria and Vella, Filippo and Zammuto, Salvatore},
  title   = {Quantum Planning for Swarm Robotics},
  journal = {Robotics and Autonomous Systems},
  volume  = {161},
  pages   = {104362},
  year    = {2023},
  doi     = {10.1016/j.robot.2023.104362}
}

@article{Hohenfeld2024QuantumNavigation,
  author  = {Hohenfeld, Hans and Heimann, Dirk and Wiebe, Felix and Kirchner, Frank},
  title   = {Quantum Deep Reinforcement Learning for Robot Navigation Tasks},
  journal = {IEEE Access},
  volume  = {12},
  pages   = {87217--87236},
  year    = {2024},
  doi     = {10.1109/ACCESS.2024.3417808}
}

@article{Mannone2025QuantumSwarm,
  author  = {Mannone, Maria and Seidita, Valeria and Chella, Antonio},
  title   = {Quantum Computing for Swarm Robotics: A Local-to-Global Approach},
  journal = {Philosophical Transactions of the Royal Society A: Mathematical, Physical and Engineering Sciences},
  volume  = {383},
  number  = {2289},
  pages   = {20240139},
  year    = {2025},
  doi     = {10.1098/rsta.2024.0139}
}

@article{Otani2025Posture,
  author  = {Otani, Takuya and Takanishi, Atsuo and Hara, Nobuyuki and Takita, Yutaka and Kimura, Koichi},
  title   = {Quantum Computation for Robot Posture Optimization},
  journal = {Scientific Reports},
  volume  = {15},
  number  = {28508},
  year    = {2025},
  doi     = {10.1038/s41598-025-12109-0}
}

@article{Zioui2021Orientation,
  author  = {Zioui, Nadjet and Mahmoudi, Yousra and Mahmoudi, Aicha and Tadjine, Mohamed and Bentouba, Said},
  title   = {A New Quantum-Computing-Based Algorithm for Robotic Arms and Rigid Bodies' Orientation},
  journal = {Journal of Applied and Computational Mechanics},
  volume  = {7},
  number  = {3},
  pages   = {1836--1846},
  year    = {2021},
  doi     = {10.22055/jacm.2021.37611.3048}
}

@article{Fazilat2022ForwardKinematics,
  author  = {Fazilat, Mehdi and Zioui, Nadjet and St-Arnaud, Jonathan},
  title   = {A Novel Quantum Model of Forward Kinematics Based on Quaternion/{P}auli Gate Equivalence: Application to a Six-Jointed Industrial Robotic Arm},
  journal = {Results in Engineering},
  volume  = {14},
  pages   = {100402},
  year    = {2022},
  doi     = {10.1016/j.rineng.2022.100402}
}

@article{Salloum2026AnnealingIK,
  author  = {Salloum, Hadi and Savin, Sergei and Kholodov, Yaroslav and Ryzhakov, Gleb and Farina, Mirko and Oseledets, Ivan},
  title   = {Quantum Annealing for Inverse Kinematics in Robotics},
  journal = {Scientific Reports},
  volume  = {16},
  number  = {4244},
  year    = {2026},
  doi     = {10.1038/s41598-025-34346-z}
}

@article{Nigatu2025QMLGrover,
  author  = {Nigatu, Hassen and Shi, Gaokun and Li, Jituo and Wang, Jin and Lu, Guodong and Li, Howard},
  title   = {Quantum Machine Learning and Grover's Algorithm for Quantum Optimization of Robotic Manipulators},
  journal = {IEEE Robotics and Automation Letters},
  volume  = {10},
  number  = {12},
  pages   = {13090--13097},
  year    = {2025},
  doi     = {10.1109/LRA.2025.3627071}
}

@article{Dahassa2025GroverArm,
  author  = {Dahassa, Mohamed S. and others},
  title   = {Optimal Control-Based Grover's Algorithm for a Six-Jointed Articulated Robotic Arm},
  journal = {Electronics},
  volume  = {14},
  number  = {13},
  pages   = {2503},
  year    = {2025},
  doi     = {10.3390/electronics14132503}
}

@article{Fazilat2025QNNIK,
  author  = {Fazilat, Mehdi and Zioui, Nadjet},
  title   = {Quantum Neural Network-Based Inverse Kinematics of a Six-Jointed Industrial Robotic Arm},
  journal = {Robotics and Autonomous Systems},
  volume  = {194},
  pages   = {105123},
  year    = {2025},
  doi     = {10.1016/j.robot.2025.105123}
}

@article{Dereli2020QPSOIK,
  author  = {Dereli, Serkan and Koker, Rasit},
  title   = {A Meta-Heuristic Proposal for Inverse Kinematics Solution of 7-{DOF} Serial Robotic Manipulator: Quantum Behaved Particle Swarm Algorithm},
  journal = {Artificial Intelligence Review},
  volume  = {53},
  number  = {2},
  pages   = {949--964},
  year    = {2020},
  doi     = {10.1007/s10462-019-09683-x}
}

@book{Nielsen2010QuantumComputation,
  author    = {Nielsen, Michael A. and Chuang, Isaac L.},
  title     = {Quantum Computation and Quantum Information},
  edition   = {10th anniversary},
  publisher = {Cambridge University Press},
  year      = {2010}
}

@article{Preskill2018NISQ,
  author  = {Preskill, John},
  title   = {Quantum Computing in the {NISQ} Era and Beyond},
  journal = {Quantum},
  volume  = {2},
  pages   = {79},
  year    = {2018},
  doi     = {10.22331/q-2018-08-06-79}
}

@article{Feynman1982Simulating,
  author  = {Feynman, Richard P.},
  title   = {Simulating Physics with Computers},
  journal = {International Journal of Theoretical Physics},
  year    = {1982},
  volume  = {21},
  number  = {6--7},
  pages   = {467--488},
  doi     = {10.1007/BF02650179}
}

@article{Lloyd1996UniversalQuantumSimulators,
  author  = {Lloyd, Seth},
  title   = {Universal Quantum Simulators},
  journal = {Science},
  year    = {1996},
  volume  = {273},
  number  = {5278},
  pages   = {1073--1078},
  doi     = {10.1126/science.273.5278.1073}
}

@inproceedings{Shor1994Algorithms,
  author    = {Shor, Peter W.},
  title     = {Algorithms for Quantum Computation: Discrete Logarithms and Factoring},
  booktitle = {Proceedings of the 35th Annual Symposium on Foundations of Computer Science},
  year      = {1994},
  pages     = {124--134},
  publisher = {IEEE Computer Society},
  doi       = {10.1109/SFCS.1994.365700}
}

@inproceedings{Grover1996Fast,
  author    = {Grover, Lov K.},
  title     = {A Fast Quantum Mechanical Algorithm for Database Search},
  booktitle = {Proceedings of the 28th Annual ACM Symposium on Theory of Computing},
  year      = {1996},
  pages     = {212--219},
  publisher = {Association for Computing Machinery},
  doi       = {10.1145/237814.237866}
}

@article{Bharti2022NISQAlgorithms,
  author  = {Bharti, Kishor and Cervera-Lierta, Alba and Kyaw, Thi Ha and Haug, Tobias and Alperin-Lea, Sumner and Anand, Abhinav and Degroote, Matthias and Heimonen, Hermanni and Kottmann, Jakob S. and Menke, Tim and Mok, Wai-Keong and Sim, Sukin and Kwek, Leong-Chuan and Aspuru-Guzik, Al{\'a}n},
  title   = {Noisy Intermediate-Scale Quantum Algorithms},
  journal = {Reviews of Modern Physics},
  year    = {2022},
  volume  = {94},
  number  = {1},
  pages   = {015004},
  doi     = {10.1103/RevModPhys.94.015004}
}

@article{Li2017HybridQuantumClassical,
  author  = {Li, Jun and Yang, Xiaodong and Peng, Xinhua and Sun, Chang-Pu},
  title   = {Hybrid Quantum-Classical Approach to Quantum Optimal Control},
  journal = {Physical Review Letters},
  volume  = {118},
  number  = {15},
  pages   = {150503},
  year    = {2017},
  doi     = {10.1103/PhysRevLett.118.150503}
}

\end{document}